\pdfoutput=1

\documentclass[11pt]{article}

\usepackage[preprint]{acl}

\usepackage{times}
\usepackage{latexsym}

\usepackage[T1]{fontenc}

\usepackage[utf8]{inputenc}

\usepackage{inconsolata}

\usepackage{graphicx}
\usepackage{amsmath}
\usepackage{amssymb}
\usepackage{booktabs}
\usepackage{multirow}
\usepackage{arydshln}
\usepackage{pgfplots}
\usepackage{fvextra} 
\DefineVerbatimEnvironment{Verbatim}{Verbatim}{breaklines=true}
\usepackage{colortbl}
\usepackage{makecell}
\usepackage{pifont}
\usepackage{todonotes}

\usepackage{listings}
\usepackage{xcolor}

\definecolor{delim}{RGB}{20,105,176}
\definecolor{numb}{RGB}{106, 109, 32}
\definecolor{string}{rgb}{0.64,0.08,0.08}

\lstdefinelanguage{json}{
    numbers=left,
    numberstyle=\small,
    frame=single,
    rulecolor=\color{black},
    showspaces=false,
    showtabs=false,
    breaklines=true,
    postbreak=\raisebox{0ex}[0ex][0ex]{\ensuremath{\color{gray}\hookrightarrow\space}},
    breakatwhitespace=true,
    basicstyle=\ttfamily\small,
    upquote=true,
    morestring=[b]",
    stringstyle=\color{string},
    literate=
     *{0}{{{\color{numb}0}}}{1}
      {1}{{{\color{numb}1}}}{1}
      {2}{{{\color{numb}2}}}{1}
      {3}{{{\color{numb}3}}}{1}
      {4}{{{\color{numb}4}}}{1}
      {5}{{{\color{numb}5}}}{1}
      {6}{{{\color{numb}6}}}{1}
      {7}{{{\color{numb}7}}}{1}
      {8}{{{\color{numb}8}}}{1}
      {9}{{{\color{numb}9}}}{1}
      {\{}{{{\color{delim}{\{}}}}{1}
      {\}}{{{\color{delim}{\}}}}}{1}
      {[}{{{\color{delim}{[}}}}{1}
      {]}{{{\color{delim}{]}}}}{1},
}

\lstdefinestyle{mypython}{
    language=Python,
    basicstyle=\ttfamily\small,
    showstringspaces=false,
    breaklines=true,
    frame=single,
    keywordstyle=\color{blue},
    stringstyle=\color{red},
    commentstyle=\color{gray},
}

\title{Schema Augmentation for Zero-Shot Domain Adaptation in Dialogue State Tracking}

\author{
    Christopher Richardson\textsuperscript{1}\thanks{Equal contribution}, 
    Roshan Sharma\textsuperscript{2}\footnotemark[1], 
    Anirudh Sundar\textsuperscript{1}\footnotemark[1], \\
    \textbf{Neeraj Gaur\textsuperscript{2}},
    \textbf{Parisa Haghani\textsuperscript{2}}, 
    \textbf{Bhuvana Ramabhadran\textsuperscript{2}} \\
    \textsuperscript{1}Georgia Institute of Technology, USA \\
    \textsuperscript{2}Google, U.S.A \\
    \texttt{crichardson8@gatech.edu} \\
}

\begin{document}
\maketitle
\begin{abstract}
Zero-shot domain adaptation for dialogue state tracking (DST) remains a challenging problem in task-oriented dialogue (TOD) systems, where models must generalize to target domains unseen at training time. Current large language model approaches for zero-shot domain adaptation rely on prompting to introduce knowledge pertaining to the target domains. However, their efficacy strongly depends on prompt engineering, as well as the zero-shot ability of the underlying language model. In this work, we devise a novel data augmentation approach, Schema Augmentation, that improves the zero-shot domain adaptation of language models. Schema Augmentation is a simple but effective technique that enhances generalization by introducing variations of slot names within the schema provided in the prompt. Experiments on MultiWOZ and SpokenWOZ showed that the proposed approach resulted in a substantial improvement over the baseline, in some experiments achieving over a twofold accuracy gain over unseen domains while maintaining equal or superior performance over all domains.
\end{abstract}

\section{Introduction}
\begin{figure}[t]
    \centering
    \includegraphics[width=0.7\columnwidth]{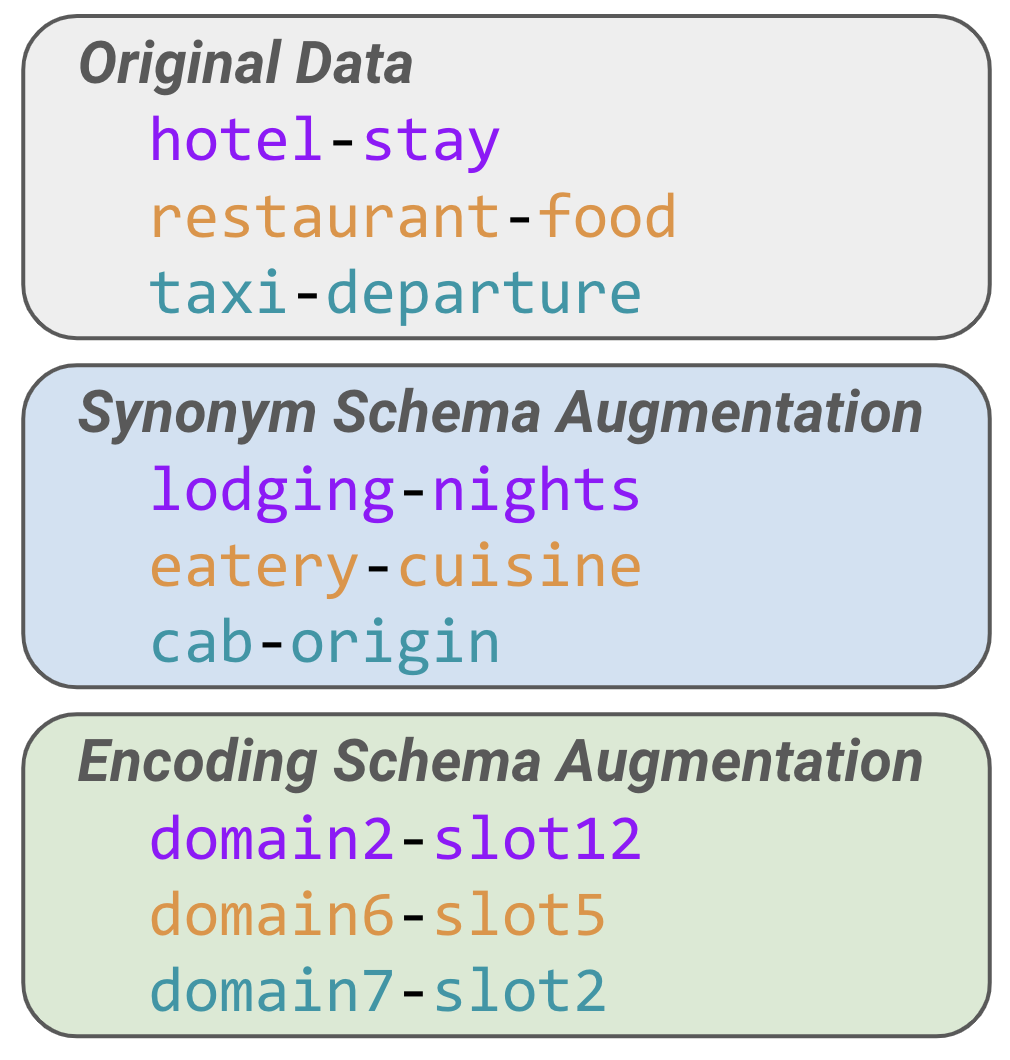}
    \caption{Examples of domain and slot replacements for both Schema Augmentation types: SSA (blue box) and ESA (green box).}
    \label{fig:example}
    \vspace{-10pt}
\end{figure}

An essential problem for task-oriented dialogue systems is Dialogue State Tracking (DST), the task of extracting a structured respresentation of the dialogue state from of user goals as the conversation progresses (see Table \ref{tab:tga_example}). While traditional DST models require extensive domain-specific annotation, instruction-tuned large language models (LLMs) enable zero-shot DST \cite{feng2023towards,yi2024survey,hosseini2020simple}. However, their effectiveness remains limited compared to specialized systems \cite{heck2023chatgpt}.

Zero-shot domain adaptation bridges this gap by allowing training on a subset of domains, improving generalization without domain-specific data \cite{aksu2023prompter,li2021zero,lin2021zero}. 
However, little work has explored end-to-end DST—modeling the full dialogue state in a single step. We address this gap by developing a novel data augmentation approach for training LLMs on zero-shot domain adaptation and evaluating it on MultiWOZ 2.1 \cite{zang2020multiwoz} and SpokenWOZ \cite{si2024spokenwoz}.

Our main contributions are:
\begin{enumerate}
\item We introduce \textit{Schema Augmentation}, a data augmentation method that improves zero-shot domain adaptation, achieving up to a twofold improvement over strong baselines.
\item We propose \textit{Target Goal Accuracy}, a new metric for evaluating domain adaptation in task-oriented dialogue.
\end{enumerate}

\section{Related Work}
\label{sec:related}
Prior work on zero-shot domain adaptation for dialogue state tracking has focused on the use of semantic slot information to generalize to unseen domains \cite{rastogi2020towards}. T5DST \cite{lin2021leveraging} utilized the slot type for more descriptive semantic slots and showed that using slot-type in conjunction with descriptions improved the zero-shot Joint Goal Accuracy. \citet{lee-etal-2021-dialogue} presented an approach that included natural language descriptions of the schema in the prompt. 

Prior work has also explored generalization in DST by reframing the task as question-answering \cite{li2021zero, heck-etal-2024-mforms}. Other approaches include parameter-efficient methods such as Prompter \cite{aksu2023prompter}, and DualLoRA \cite{luo2024zero}. Prompter applied prompt-based learning with in-context examples, while DualLoRA separated the adapters used for the prompts and the dialogue context to assist generalizability. 
D3ST \cite{zhao2022description} relied solely on schema descriptions and represents the current state of the art in zero-shot domain adaptation on MultiWOZ 2.1. In contrast to D3ST that replaces the schema, this work focuses on \textit{augmenting} the dataset with new samples with modified schemata.  Prior work has provided preliminary evidence on the benefits of data augmentation \cite{ma2019end} by back-translating sentences between English and Chinese. This work formalizes the schema augmentation using synonym augmentation from a controlled set of synonyms and an encoding-based schema augmentation that relies on random replacement. 

\section{Problem Formulation}
Dialogue states are constructed from sets of slots $\mathcal{S}$ and values $\mathcal{V}$, with each slot belonging to a particular set of domains $\mathcal{D}$. A dataset $\mathcal{X}=\{(x,y)\}$ consists of dialogues $x$ and dialogue states $y$. A dialogue state is a set of $K$ distinct slot/value pairs, where $K \geq 0$ and differs for every $(x,y)$:
\begin{align}
    y = \{(s_k,v_k)\}_{k=1}^K: s\in\mathcal{S}, v\in\mathcal{V}
\end{align}
The goal is to learn a function that maps dialogue to dialogue state, $\pi(x) = y$. When evaluating domain adaptation, we are particularly interested in predicting slots that belong to our target domains, which we will call $\mathcal{D}_T$. Thus the dialogue \textit{substate} of interest $y_T \subseteq y$ is:
\begin{align}
    y_T \overset{\Delta}{=} \{(s,v)\}: s\in\mathcal{D}_T
\end{align}
Similarly, we denote the target-domain subset of the predicted state as $\pi_T(x)$. To most accurately reflect true domain adaptation performance, we look at only those dialogues that have nonempty target-domain substates:
\begin{align}
    \mathcal{X}_T = \{(x,y_T)\} : y_T \neq \varnothing
\end{align}


For each slot $s\in\mathcal{S}$ we are given a description of the slot and a list of possible values it can take. This information, referred to as the \textit{schema}, defines the structure of slot names, descriptions, and possible values, and is provided in the prompt.

\section{Methods}

\newcommand{\xmark}{\ding{55}}

\subsection{Target Goal Accuracy} \label{sec:tga}

The primary metric used in the DST literature is Joint Goal Accuracy (JGA) \cite{henderson2014second}, which looks at all slots and domains, and thus does not measure domain adaptation performance directly. To alleviate this, we introduce a new metric: Target Goal Accuracy (TGA), a sub-goal version of JGA that considers only those slots belonging to the target domains. 

\paragraph{Joint Goal Accuracy (JGA):} The fraction of turns in which the entire state is accurately predicted. We include all dialogue states, including empty states, to be comprehensive.
\paragraph{Target Goal Accuracy (TGA):} The fraction of all target-domain turns (turns with at least one slot belonging to the target domains) in which the target-domain substate is accurately predicted. We do not include turns that have no target-domain slots active in the ground truth dialogue state, as these empty states would dominate the overall metric.

Formally, we define these metrics in terms of hits over a dataset $\mathcal{X}$, i.e samples for which the model predicted the correct overall state:
\begin{gather}
    \mathcal{H} = \{(x,y) \in \mathcal{X} : \; \pi(x) = y \} \\
    \text{JGA} = \frac{|\mathcal{H}|}{|\mathcal{X}|} 
\end{gather}
TGA only considers the target subset:
\begin{gather}
    \mathcal{H}_T = \{(x,y_T) \in \mathcal{X}_T : \; \pi_T(x) = y_T \} \\
    \text{TGA} = \frac{|\mathcal{H}_T|}{|\mathcal{X}_T|} 
\end{gather}
JGA gives a standardized measure of overall performance on DST, while TGA highlights domain adaptation abilities of each method. Our goal is to improve TGA without degradation of JGA. 

\begin{table*}
\scriptsize
    \centering
    \begin{tabular}{clllcc}
    \toprule
    \textbf{Turn} & \textbf{Dialogue} & \textbf{Dialogue State} & \textbf{Model Response} & \textbf{JGA} & \textbf{TGA} \\
    \midrule
      & \textbf{USER:} Hello I'd like to book a table for Friday. & \texttt{restaurant-day: friday} & \texttt{\textcolor{blue}{restaurant-day: friday}} && \\
    1 & \textbf{SYSTEM:} I see openings at Chotchkie's all evening, is that ok? & \texttt{restaurant-time: -} & \texttt{\textcolor{blue}{restaurant-time: -}} & \checkmark & N/A \\
      & & \texttt{taxi-arriveby: -} & \texttt{\textcolor{blue}{taxi-arriveby: -}} && \\
    \rowcolor{gray!20} 
      & \textbf{USER:} Yes, let's do 19:00. & \texttt{restaurant-day: friday} & \texttt{\textcolor{blue}{restaurant-day: friday}} && \\
    \rowcolor{gray!20} 
    2 & \textbf{SYSTEM:} Great, it's booked. Can I help you with anything else? & \texttt{restaurant-time: 19:00} & \texttt{\textcolor{blue}{restaurant-time: 19:00}} & \checkmark & N/A\\
    \rowcolor{gray!20} 
      & & \texttt{taxi-arriveby: -} & \texttt{\textcolor{blue}{taxi-arriveby: -}} && \\
      & \textbf{USER:} I also need a taxi to pick me up 30 minutes before. & \texttt{restaurant-day: friday} & \texttt{\textcolor{blue}{restaurant-day: friday}} &&\\
    3 & \textbf{SYSTEM:} Okay, I scheduled it. & \texttt{restaurant-time: 19:00} & \texttt{\textcolor{blue}{restaurant-time: 19:00}} & \xmark & \xmark \\
      & & \texttt{taxi-arriveby: 18:30} & \texttt{\textcolor{red}{taxi-arriveby: -}} && \\
    \midrule
    & & & \multicolumn{1}{r}{\textbf{Total}} & \textbf{67\%} & \textbf{0\%} \\ 
    \bottomrule
    \end{tabular}
    \caption{Example dialogue illustrating the difference between JGA and TGA for the \texttt{taxi} domain. The last two columns show whether the response triggers a hit or miss for the correpsponding metric, and N/A means that example will not be counted in the metric. TGA better reflects performance on \texttt{taxi} by ignoring samples with empty \texttt{taxi} substates. A model that never attempts to generate \texttt{taxi} slots can potentially still achieve a high JGA.}
    \vspace{-10pt}
    \label{tab:tga_example}
\end{table*}

We are motivated by the limitations of prior work in measuring true domain adaptation, which follows a "leave-one-out" cross-domain adaptation setup \cite{lin2021zero, lin2021leveraging}. In this setup, one domain is withheld from training, and the model is trained on the remaining domains before being tested on the holdout domain. We argue that this does not capture true domain adaptation due to two key issues: \textit{slot overlap} and \textit{empty substate bias}.

\textit{Slot overlap} occurs when slots are shared across domains, reducing the challenge of encountering unseen slots. Table \ref{tab:slots_mwoz} shows that domains like taxi/train/bus and hotel/restaurant have significant slot overlap, making adaptation easier when similar slots exist in training.

The \textit{empty substate bias} arises from how prior work evaluates zero-shot domain performance using JGA. The test set is filtered to include only samples labeled with the target domain, and only target domain slots are measured for JGA calculation. However, prior work does not filter out empty slots, leading to cases where the entire substate for the target domain is empty. This is especially problematic for domains appearing later in dialogues, such as \texttt{taxi}, resulting in artificially inflated JGA scores.

Table \ref{tab:tga_example} illustrates this issue with a MultiWOZ-style example. A model with no taxi training never fills taxi slots. When evaluating JGA for taxi, turns 1-2 count as correct because the model "predicted" the empty taxi substate—despite lacking any knowledge of taxi. In real-world scenarios, such "hits" are irrelevant, so we propose TGA to better reflect true domain adaptation. In this example, JGA is 2/3 (\textbf{67\%}), while TGA is 0/1 (\textbf{0\%}).

\subsection{Schema Augmentation}

In our method, \emph{Schema Augmentation}, we generate augmented data by altering domain and slot names to enhance the model’s robustness to schema changes. We explore two types of Schema Augmentation: \textit{Synonym Schema Augmentation (SSA)}, in which we replace domain and slot names with synonyms; and \textit{Encoding Schema Augmentation (ESA)}, in which we replace domains and slots with non-semantic codes, requiring the model to rely on slot descriptions and possible values to predict the dialogue state accurately. Figure \ref{fig:example} illustrates and example of these two types of replacements.

For both types of augmentation, we utilize multiple possible replacements for each slot and domain. We randomly select the replacement for each sample from a pre-determined list. Gemini \cite{team2023gemini} provides the synonym lists, while we create encodings by appending integers to ‘domain’ or ‘slot.’ In ESA, we ensure non-overlapping encodings for each domain and slot by using disjoint ranges of integers. Appendix \ref{sec:appendix} describes replacements used in both cases.

\section{Experiments}

\subsection{Modeling}
We perform dialogue state tracking end-to-end by fine-tuning \texttt{gemma-2-9b-it} \cite{team2024gemma} on prompts constructed from dialogue data paired with dialogue state outputs. We build on LDST \cite{feng2023towards} for our prompt construction. Our prompts include an instruction, DST schema, and dialogue, and outputs are represented as textual JSON. Details of our experimental hyperparameters and compute are included in Appendix \ref{sec:appendix}, as well as full examples of our prompts.

All of our experiments use Gemma \cite{team2024gemma}. We utilize the \texttt{gemma-2-9b-it} variant as it is feasible to train on a single GPU, and we use the instruction tuned version in order to enable understanding of language feedback.

\subsection{Datasets and Baselines}
We conduct our experiments on two open-source dialogue state tracking datasets --- \textbf{MultiWOZ2.1} \cite{eric2019multiwoz} and \textbf{SpokenWOZ}  \cite{si2024spokenwoz}, a spoken TOD dataset inspired by MultiWOZ. For ablation studies, we perform experiments on MultiWOZ 2.2 \cite{zang2020multiwoz}, a cleaner update to 2.1 with less noise. A detailed description is provided in \ref{sec:appendix_dataset_description}. 

In order to best simulate the domain adaptation scenario, we select holdout domains that minimize slot overlap with the training domains. To this end, we choose $\mathcal{D}_T=\texttt{\{taxi,train\}}$ as our holdout domain set. Tables showing slot overlap are included in Appendix \ref{sec:appendix} (Table \ref{tab:slots_mwoz}).

We compare our approach against several baselines from prior work on zero-shot dialogue state tracking and cross-domain transfer --- \textbf{TransferQA}, \textbf{T5DST}, \textbf{Prompter}, \textbf{DualLoRA}, and \textbf{D3ST}. The methods are described in Section \ref{sec:related}.

\definecolor{lightlightgray}{gray}{0.9} 
\begin{table}[t]
    \centering
    \begin{tabular}{p{2.7cm}ccc}
        \toprule
        \textbf{Method} & $\textbf{JGA}_{\text{taxi}}$ & $\textbf{JGA}_{\text{train}}$ & \textbf{TGA} \\
        \midrule
        \rowcolor{lightlightgray} \multicolumn{4}{c}{\textbf{\textit{Prior work}}} \\
        TransferQA & 61.9 & 36.7 & 15.9 \\
        T5DST & 64.6 & 35.4 & 15.7 \\
        Prompter & 66.3 & 39.0 & 20.2 \\
        DualLoRA & 67.2 & 42.4 & 24.2 \\
        D3ST & \textbf{78.4}	& 38.7 & 25.2 \\ 
        \midrule
        \rowcolor{lightlightgray} \multicolumn{4}{c}{\textbf{\textit{Our approach}}} \\
        SSA & 66.7 & \textit{48.8} & \textit{31.8} \\
        ESA & \textit{69.5} & \textbf{50.6} & \textbf{34.9} \\
        	
        \bottomrule
    \end{tabular}
    \caption{Comparison of our method to prior work on MultiWOZ 2.1. Goal accuracies (\%) showing JGA reported in original prior works with the corresponding estimated TGA.}
    \vspace{-10pt}
    \label{tab:mwoz21}
\end{table}
\footnotetext[1]{Our reimplementation of \cite{zhao2022description}.}




\subsection{Main Results}
Table \ref{tab:mwoz21} shows results on the MultiWOZ 2.1 dataset alongside results from prior work. For direct comparison to our method, we only report results for JGA on \texttt{taxi} and \texttt{train} domains, as well as TGA. Details of computation of TGA from JGA for prior work can be found in the Appendix \ref{sec:computing_tga}. Our method shows an improvement of 19.6\% TGA (absolute) over the best performing baseline, D3ST \cite{zhao2022description}, despite using a model with 20\% fewer parameters (9B vs. 11B). There are a few things to note about approaches in prior work that confer an advantage. For one, they do not address the issue of slot overlap. Previous experiment using the leave-one-out approach ignore the \textit{slot overlap} problem (see Section \ref{sec:tga} for reference) and train models on many of the same slots that are seen in the target domain. This gives a large advantage, particularly when evaluating on domains like \texttt{taxi} whose slots are all included in domains (\texttt{train}) present in the training data.

Our main results are shown in Table \ref{tab:mwoz21}. Our best method, ESA, improves TGA over the best performing baseline, D3ST, by 9.7\%. It also improves JGA on \texttt{train} by 11.9\% and achieves the second highest JGA on \texttt{taxi}, despite having no access to those slots during training time, unlike all the baselines we compare to.

We also report results using \texttt{Mistral-7B-Instruct-v0.3} \cite{jiang2023mistral}, which is shown in the Appendix.

\subsection{SpokenWOZ Results}
\definecolor{lightlightgray}{gray}{0.9} 
\begin{table}[t]
    \centering
    \begin{tabular}{p{2.5cm}ccc}
        \toprule
        \textbf{Method} & $\textbf{JGA}_{\text{taxi}}$ & $\textbf{JGA}_{\text{train}}$ & \textbf{TGA} \\
        \midrule
        \rowcolor{lightlightgray} \multicolumn{4}{c}{\textbf{\textit{Prior work}}} \\
        D3ST \footnotemark[1] & \textit{67.1} & 41.1 & 21.9 \\ 
        \rowcolor{lightlightgray} \multicolumn{4}{c}{\textbf{\textit{Our approach}}} \\
        SSA & \textbf{67.6} & \textit{49.6} & \textit{28.8} \\
        ESA & 66.7 & \textbf{50.5} & \textbf{30.3}  \\
        \bottomrule
    \end{tabular}
    \caption{SpokenWOZ results. Comparison of our method to the prior state-of-the-art D3ST. Note that JGA here is overall JGA on the full test set with all domains.}
    \label{tab:swoz}
\end{table}

To further assess our method, we also run experiments on SpokenWOZ. 
To the best of our knowledge, no prior work on cross-domain transfer has been done on SpokenWOZ, so we re-implement the highest performing baseline from MultiWOZ (D3ST, \citet{zhao2022description}) for comparison. The results are shown in Table \ref{tab:swoz}, with our method outperforming D3ST.

\section{Conclusion}

We developed Schema Augmentation, a data augmentation technique for zero-shot domain adaptation in dialogue state tracking. To assess its effectiveness, we introduced Target Goal Accuracy (TGA), a metric that evaluates performance specifically on unseen target domains. Our method significantly boosts TGA compared to baseline approaches across two widely-used datasets, demonstrating robustness to out-of-distribution information in prompts at inference. 

\section*{Limitations}
Our work has several limitations. We evaluated Schema Augmentation on two instruction-tuned models, Gemma and Mistral, but did not explore a broader range of models, so the generalizability of our findings to other architectures is unclear. Additionally, while we tested on two task-oriented dialogue datasets (MultiWOZ and SpokenWOZ), both are based on similar domains, and further testing on more diverse datasets is needed. Additionally, we only used one set of holdout domains, whereas our experiments could be repeated using different sets of the available domains as holdouts. Due to compute constraints, we limited fine-tuning to models with fewer than 10 billion parameters, which may affect performance compared to larger models. Moreover, our experiments were confined to English-language datasets, leaving the effectiveness of Schema Augmentation in multilingual or non-English contexts unexplored. Lastly, the scope of hyperparameter tuning was limited by available resources, and further exploration of fine-tuning configurations could yield even more insights.

\section*{Ethics Statement}
This work aims to improve dialogue state tracking in task-oriented systems, with potential applications in real-world settings like customer service or healthcare. Ensuring the fairness and robustness of these models is crucial to avoid biased or harmful outcomes, especially for underrepresented groups. Additionally, while our method enhances model performance in unseen domains, careful consideration is required before deploying such models in sensitive areas where errors could have significant consequences. Finally, the environmental impact of training large models is an important factor, and more sustainable practices in AI research should be prioritized.

\bibliography{main}

\begin{thebibliography}{22}
\providecommand{\natexlab}[1]{#1}

\bibitem[{Aksu et~al.(2023)Aksu, Kan, and Chen}]{aksu2023prompter}
Taha Aksu, Min-Yen Kan, and Nancy~F Chen. 2023.
\newblock Prompter: Zero-shot adaptive prefixes for dialogue state tracking domain adaptation.
\newblock \emph{arXiv preprint arXiv:2306.04724}.

\bibitem[{Eric et~al.(2019)Eric, Goel, Paul, Kumar, Sethi, Ku, Goyal, Agarwal, Gao, and Hakkani-Tur}]{eric2019multiwoz}
Mihail Eric, Rahul Goel, Shachi Paul, Adarsh Kumar, Abhishek Sethi, Peter Ku, Anuj~Kumar Goyal, Sanchit Agarwal, Shuyang Gao, and Dilek Hakkani-Tur. 2019.
\newblock Multiwoz 2.1: A consolidated multi-domain dialogue dataset with state corrections and state tracking baselines.
\newblock \emph{arXiv preprint arXiv:1907.01669}.

\bibitem[{Feng et~al.(2023)Feng, Lu, Liu, Zhan, and Wu}]{feng2023towards}
Yujie Feng, Zexin Lu, Bo~Liu, Liming Zhan, and Xiao-Ming Wu. 2023.
\newblock Towards llm-driven dialogue state tracking.
\newblock \emph{arXiv preprint arXiv:2310.14970}.

\bibitem[{{Gemini Team}(2023)}]{team2023gemini}
{Gemini Team}. 2023.
\newblock Gemini: a family of highly capable multimodal models.
\newblock \emph{arXiv preprint arXiv:2312.11805}.

\bibitem[{{Gemma Team}(2024)}]{team2024gemma}
{Gemma Team}. 2024.
\newblock Gemma: Open models based on gemini research and technology.
\newblock \emph{arXiv preprint arXiv:2403.08295}.

\bibitem[{Heck et~al.(2024)Heck, Heck, and Sundar}]{heck-etal-2024-mforms}
Larry Heck, Simon Heck, and Anirudh~S. Sundar. 2024.
\newblock \href {https://aclanthology.org/2024.lrec-main.984/} {m{F}orms : Multimodal form filling with question answering}.
\newblock In \emph{Proceedings of the 2024 Joint International Conference on Computational Linguistics, Language Resources and Evaluation (LREC-COLING 2024)}, pages 11262--11271, Torino, Italia. ELRA and ICCL.

\bibitem[{Heck et~al.(2023)Heck, Lubis, Ruppik, Vukovic, Feng, Geishauser, Lin, van Niekerk, and Ga{\v{s}}i{\'c}}]{heck2023chatgpt}
Michael Heck, Nurul Lubis, Benjamin Ruppik, Renato Vukovic, Shutong Feng, Christian Geishauser, Hsien-Chin Lin, Carel van Niekerk, and Milica Ga{\v{s}}i{\'c}. 2023.
\newblock Chatgpt for zero-shot dialogue state tracking: A solution or an opportunity?
\newblock \emph{arXiv preprint arXiv:2306.01386}.

\bibitem[{Henderson et~al.(2014)Henderson, Thomson, and Williams}]{henderson2014second}
Matthew Henderson, Blaise Thomson, and Jason~D Williams. 2014.
\newblock The second dialog state tracking challenge.
\newblock In \emph{Proceedings of the 15th annual meeting of the special interest group on discourse and dialogue (SIGDIAL)}, pages 263--272.

\bibitem[{Hosseini-Asl et~al.(2020)Hosseini-Asl, McCann, Wu, Yavuz, and Socher}]{hosseini2020simple}
Ehsan Hosseini-Asl, Bryan McCann, Chien-Sheng Wu, Semih Yavuz, and Richard Socher. 2020.
\newblock A simple language model for task-oriented dialogue.
\newblock \emph{Advances in Neural Information Processing Systems}, 33:20179--20191.

\bibitem[{Hu et~al.(2021)Hu, Shen, Wallis, Allen-Zhu, Li, Wang, Wang, and Chen}]{hu2021lora}
Edward~J Hu, Yelong Shen, Phillip Wallis, Zeyuan Allen-Zhu, Yuanzhi Li, Shean Wang, Lu~Wang, and Weizhu Chen. 2021.
\newblock Lora: Low-rank adaptation of large language models.
\newblock \emph{arXiv preprint arXiv:2106.09685}.

\bibitem[{Jiang et~al.(2023)Jiang, Sablayrolles, Mensch, Bamford, Chaplot, Casas, Bressand, Lengyel, Lample, Saulnier et~al.}]{jiang2023mistral}
Albert~Q Jiang, Alexandre Sablayrolles, Arthur Mensch, Chris Bamford, Devendra~Singh Chaplot, Diego de~las Casas, Florian Bressand, Gianna Lengyel, Guillaume Lample, Lucile Saulnier, et~al. 2023.
\newblock Mistral 7b.
\newblock \emph{arXiv preprint arXiv:2310.06825}.

\bibitem[{Lee et~al.(2021)Lee, Cheng, and Ostendorf}]{lee-etal-2021-dialogue}
Chia-Hsuan Lee, Hao Cheng, and Mari Ostendorf. 2021.
\newblock \href {https://doi.org/10.18653/v1/2021.emnlp-main.404} {Dialogue state tracking with a language model using schema-driven prompting}.
\newblock In \emph{Proceedings of the 2021 Conference on Empirical Methods in Natural Language Processing}, pages 4937--4949, Online and Punta Cana, Dominican Republic. Association for Computational Linguistics.

\bibitem[{Li et~al.(2021)Li, Cao, Sridhar, Zhu, Li, Hamza, and McAuley}]{li2021zero}
Shuyang Li, Jin Cao, Mukund Sridhar, Henghui Zhu, Shang-Wen Li, Wael Hamza, and Julian McAuley. 2021.
\newblock Zero-shot generalization in dialog state tracking through generative question answering.
\newblock \emph{arXiv preprint arXiv:2101.08333}.

\bibitem[{Lin et~al.(2021{\natexlab{a}})Lin, Liu, Madotto, Moon, Crook, Zhou, Wang, Yu, Cho, Subba et~al.}]{lin2021zero}
Zhaojiang Lin, Bing Liu, Andrea Madotto, Seungwhan Moon, Paul Crook, Zhenpeng Zhou, Zhiguang Wang, Zhou Yu, Eunjoon Cho, Rajen Subba, et~al. 2021{\natexlab{a}}.
\newblock Zero-shot dialogue state tracking via cross-task transfer.
\newblock \emph{arXiv preprint arXiv:2109.04655}.

\bibitem[{Lin et~al.(2021{\natexlab{b}})Lin, Liu, Moon, Crook, Zhou, Wang, Yu, Madotto, Cho, and Subba}]{lin2021leveraging}
Zhaojiang Lin, Bing Liu, Seungwhan Moon, Paul Crook, Zhenpeng Zhou, Zhiguang Wang, Zhou Yu, Andrea Madotto, Eunjoon Cho, and Rajen Subba. 2021{\natexlab{b}}.
\newblock Leveraging slot descriptions for zero-shot cross-domain dialogue state tracking.
\newblock \emph{arXiv preprint arXiv:2105.04222}.

\bibitem[{Luo et~al.(2024)Luo, Tang, Wang, and Zhang}]{luo2024zero}
Xiang Luo, Zhiwen Tang, Jin Wang, and Xuejie Zhang. 2024.
\newblock Zero-shot cross-domain dialogue state tracking via dual low-rank adaptation.
\newblock \emph{arXiv preprint arXiv:2407.21633}.

\bibitem[{Ma et~al.(2019)Ma, Zeng, Zhu, Li, Yang, Yao, Zhou, and Shen}]{ma2019end}
Yue Ma, Zengfeng Zeng, Dawei Zhu, Xuan Li, Yiying Yang, Xiaoyuan Yao, Kaijie Zhou, and Jianping Shen. 2019.
\newblock An end-to-end dialogue state tracking system with machine reading comprehension and wide \& deep classification.
\newblock \emph{arXiv preprint arXiv:1912.09297}.

\bibitem[{Rastogi et~al.(2020)Rastogi, Zang, Sunkara, Gupta, and Khaitan}]{rastogi2020towards}
Abhinav Rastogi, Xiaoxue Zang, Srinivas Sunkara, Raghav Gupta, and Pranav Khaitan. 2020.
\newblock Towards scalable multi-domain conversational agents: The schema-guided dialogue dataset.
\newblock In \emph{Proceedings of the AAAI conference on artificial intelligence}, volume~34, pages 8689--8696.

\bibitem[{Si et~al.(2024)Si, Ma, Gao, Wu, Lin, Dai, Li, Yan, Huang, and Li}]{si2024spokenwoz}
Shuzheng Si, Wentao Ma, Haoyu Gao, Yuchuan Wu, Ting-En Lin, Yinpei Dai, Hangyu Li, Rui Yan, Fei Huang, and Yongbin Li. 2024.
\newblock Spokenwoz: A large-scale speech-text benchmark for spoken task-oriented dialogue agents.
\newblock \emph{Advances in Neural Information Processing Systems}, 36.

\bibitem[{Yi et~al.(2024)Yi, Ouyang, Liu, Liao, Xu, and Shen}]{yi2024survey}
Zihao Yi, Jiarui Ouyang, Yuwen Liu, Tianhao Liao, Zhe Xu, and Ying Shen. 2024.
\newblock A survey on recent advances in llm-based multi-turn dialogue systems.
\newblock \emph{arXiv preprint arXiv:2402.18013}.

\bibitem[{Zang et~al.(2020)Zang, Rastogi, Sunkara, Gupta, Zhang, and Chen}]{zang2020multiwoz}
Xiaoxue Zang, Abhinav Rastogi, Srinivas Sunkara, Raghav Gupta, Jianguo Zhang, and Jindong Chen. 2020.
\newblock Multiwoz 2.2: A dialogue dataset with additional annotation corrections and state tracking baselines.
\newblock \emph{arXiv preprint arXiv:2007.12720}.

\bibitem[{Zhao et~al.(2022)Zhao, Gupta, Cao, Yu, Wang, Lee, Rastogi, Shafran, and Wu}]{zhao2022description}
Jeffrey Zhao, Raghav Gupta, Yuan Cao, Dian Yu, Mingqiu Wang, Harrison Lee, Abhinav Rastogi, Izhak Shafran, and Yonghui Wu. 2022.
\newblock Description-driven task-oriented dialog modeling.
\newblock \emph{arXiv preprint arXiv:2201.08904}.

\end{thebibliography}

\appendix

\section{Appendix} \label{sec:appendix}
\subsection{Ablation Study}
\begin{figure}[t]
    \centering
    \begin{tikzpicture}
    \begin{axis}[
    	x tick label style={
    		/pgf/number format/1000 sep=},
    	xlabel=TGA (\%) vs. Experiment,
    	xlabel style={at={(axis description cs:0.5,1.1)},anchor=south},
    	enlargelimits=0.2,
    	legend style={at={(0.5,-0.15)},
    	anchor=north,legend columns=-1},
    	ybar,
            bar width=20pt,
    	xmin=1.9, xmax=4.1,
    	xtick={2,3,4},
    	xticklabels={Baseline(no aug.), SSA, ESA},
        nodes near coords, 
        nodes near coords align={vertical}, 
        every node near coord/.append style={font=\small} 
    ]
    \addplot 
    	coordinates {(2,18.2) (3,34.5) (4,40.7)};
    \addplot 
    	coordinates {(2,10.4) (3,20.0) (4,16.5)};
    \legend{Original,Shuffled Descriptions/Values}
    \end{axis}
    \end{tikzpicture}
    \caption{Ablation results for MultiWOZ 2.2. SSA and ESA are our Schema Augmentation methods. No Aug means fine-tuning \texttt{gemma-2-9b-it} without data augmentation, to illustrate the effects of Schema Augmentation.}
    \label{fig:ablation}
\end{figure}

We conduct an ablation study to shed light on the mechanism of Schema Augmentation, and why the encoding variant generally outperforms the synonym variant. In our ablation, we introduce a mismatch between slots and their corresponding descriptions/values in the schema by random shuffling, i.e. every slot is paired with a randomly chosen description and value list from a different slot during preprocessing. The shuffling is randomized on every sample to ensure there is no correlation between description/values and the correct slot in the answer. We conduct our study on the MultiWOZ 2.2 dataset, which was preferred over 2.1 due to its higher quality and lower noise content. Figure \ref{fig:ablation} shows the ablation results. We compared our methods SSA and ESA to a baseline of standard fine-tuning without data augmentation, but using the same hyperparameters and model (\texttt{gemma-2-9b-it}). Upon shuffling descriptions/values, we observe the greatest drop in TGA for ESA (-24.5\%), and second greatest for SSA (-14.5\%). The baseline had the lowest drop (-7.8\%). A higher drop in TGA means a greater desgradation in performance when the information content of descriptions/values is effectively removed. This suggests that Schema Augmentation encourages the model to pay more attention to the descriptions and values of the slots, and that effect is stronger with ESA than SSA. 

\subsection{Computing TGA} \label{sec:computing_tga}
Prior work uses the "leave-one-out" cross-domain adaptation setup \cite{lin2021zero, lin2021leveraging}. In this scenario, a single domain at a time is designated as the holdout domain $d_H$ and withheld from the training set. A model is trained on the remaining four domains, and then tested on the test split. Prior work reports JGA for $d_H$, computed over dialogues that include $d_H$. We will refer to this as holdout JGA, or $\text{JGA}^H$.

In our experiments, we designated \texttt{taxi} and \texttt{train} as holdout domains due to slot overlap (see Section \ref{sec:tga}). In order to convert $\text{JGA}^H$ to TGA, we will combine $\text{JGA}^{\texttt{taxi}}$ and $\text{JGA}^{\texttt{train}}$ using the known test set distribution over domains:
\begin{align*}
    \text{JGA}^{\texttt{taxi}+\texttt{train}} = \text{JGA}^{\texttt{taxi}} \cdot N_{\texttt{taxi}} \\
                                                + \text{JGA}^{\texttt{train}} \cdot N_{\texttt{train}}
\end{align*}
We also need to recognize that unlike TGA, $\text{JGA}^H$ includes empty states. This is due to the fact that the domain filtering is done \textit{per dialogue}, but the data itself is \textit{per turn}. For example, a dialogue labelled with the \texttt{taxi} domain can have many turns which have empty \texttt{taxi} slots - all the turns which appear before \texttt{taxi} is metioned in the dialogue. TGA is a stricter metric that does not consider these empty states. In order to convert to TGA, we thus need an additional assumption about the performance on empty states. In all experiments done in this work, the accuracy of empty states was >99\% on all domains for both datasets. Thus, without a significant loss in accuracy, we assume a 100\% accuracy of prior work on empty states. This will not be perfectly accurate, but without generations provided by prior work, we cannot perfectly determine this accuracy. With this assumption, we can now estimate TGA:
\begin{align}
    \text{TGA}^{\texttt{taxi},\text{est}} = \frac{\text{JGA}^{\texttt{taxi}} \cdot N_{\texttt{taxi}} - N_{\texttt{taxi}}^{\text{empty}}}{N_{\texttt{taxi}}^{\text{nonempty}}}
\end{align}
The formula for $\text{TGA}^{\texttt{train},est}$ is the same, and we combine them to get our overall estimate of TGA:
\begin{align}
    \text{TGA}^{\text{est}} &= \text{TGA}^{\texttt{taxi},\text{est}} \cdot N_{\texttt{taxi}}^{\text{nonempty}} \nonumber \\
                            &+ \text{TGA}^{\texttt{train},\text{est}} \cdot N_{\texttt{train}}^{\text{nonempty}} \label{eqn:tga}
\end{align}

\subsection{Dataset Description}
\label{sec:appendix_dataset_description}
MultiWOZ is a multi-domain task-oriented dialogue dataset comprising annotated dialogues across eight domains including hotel booking, restaurant reservation, and taxi ordering. Each dialogue is annotated with the dialogue state at each turn. For consistency with prior work, we run our main experiments on MultiWOZ 2.1 \cite{eric2019multiwoz}, a popular revision of the original. We also report results on SpokenWOZ \cite{si2024spokenwoz}. SpokenWOZ dialogues were collected from crowdworkers engaging in spoken conversations and includes text transcriptions from an automatic speech recognition (ASR) system. We perform our experiments on the audio transcriptions. For ablation studies, we use a cleaner version of MultiWOZ, MultiWOZ 2.2 \cite{zang2020multiwoz} due to the lower noise content.

\subsection{Hyperparameters}
For our experiments, we fine-tune both our models using the AdamW optimizer with a learning rate of 2e-4 and warmup ratio of 0.03. Due to compute constraints, we use LoRA \cite{hu2021lora} to train adapters while keeping base weights frozen. We use a LoRA $r=2$ and $\alpha=2$ with a dropout of 0, and adapter weights added to all linear layers. In all experiments, the modes is fine-tuned to convergence in each experiment. We achieve this by evaluating on the validation split each epoch and choosing an early stopping patience of 1. This ensures that each experiment yields the best model and comparisons between methods are fair.

All experiments use a random seed of 42 and deterministic algorithms were used everywhere possible to ensure minimal variation between runs. All accuracy metrics reported had less than 1\% variance across all runs.

\subsection{Mistral Results}

Our results on the Mistral 7B model from Table \ref{tab:mistral_results} corroborate our findings that the proposed augmentation methods yield significant gains for dialog state tracking when applied to multiple different LLMs. 

\definecolor{lightlightgray}{gray}{0.9} 
\begin{table*}[h]
    \centering
    \begin{tabular}{p{2.5cm}ccc}
        \toprule
        \textbf{Method} & $\textbf{JGA}_{\text{taxi}}$ & $\textbf{JGA}_{\text{train}}$ & \textbf{TGA} \\
        \midrule
        \rowcolor{lightlightgray} \multicolumn{4}{c}{\textbf{\textit{MultiWOZ 2.2}}} \\
        SSA & 65.0 & 42.9 & 24.9 \\ 
        ESA & 67.9 & 42.6 & 26.0 \\
        \rowcolor{lightlightgray} \multicolumn{4}{c}{\textbf{\textit{SpokenWOZ}}} \\
        SSA & 66.7 & 35.2 & 12.5 \\
        ESA & 66.7 & 37.7 & 15.4  \\
        \bottomrule
    \end{tabular}
    \caption{MultiWOZ 2.2 and SpokenWOZ results using the Huggingface implementation of the \texttt{Mistral-7B-Instruct-v0.3} model.}
    \label{tab:mistral_results}
\end{table*}

\subsection{Compute}
All fine-tuning and inference was run on Nvidia A40 GPUs with 48GB GDDR6 memory. Fine-tuning took ~1-2 hours on 8 GPUs in parallel with pytorch distributed data parallel (DDP).

\subsection{Slots and Domains}
Tables \ref{tab:slots_mwoz} and \ref{tab:slots_swoz} show the domain and slot combinations for the two datasets. Taxi, train, and bus were chosen as holdout domains due to many slots in common with each other and few slots in common with other domains. 

\begin{table*}
\centering
    \begin{tabular}{lccccccc}
    \toprule
     & \textbf{attraction} & \textbf{hotel} & \textbf{restaurant} & \textbf{taxi} & \textbf{train} \\
    \midrule
    area & \checkmark & \checkmark & \checkmark &  &  \\
    arriveby &  &  &  & \checkmark & \checkmark \\
    bookday &  & \checkmark & \checkmark &  &  \\
    bookpeople &  & \checkmark & \checkmark &  & \checkmark \\
    bookstay &  & \checkmark &  &  &  \\
    booktime &  &  & \checkmark &  &  \\
    day &  &  &  &  & \checkmark \\
    departure &  &  &  & \checkmark & \checkmark \\
    destination &  &  &  & \checkmark & \checkmark \\
    food &  &  & \checkmark &  &  \\
    internet &  & \checkmark &  &  &  \\
    leaveat &  &  &  & \checkmark & \checkmark \\
    name & \checkmark & \checkmark & \checkmark &  &  \\
    parking &  & \checkmark &  &  &  \\
    pricerange &  & \checkmark & \checkmark &  &  \\
    stars &  & \checkmark &  &  &  \\
    type & \checkmark & \checkmark &  &  & \\
    \bottomrule
    \end{tabular}
\caption{Domain-Slot Combinations for MultiWOZ.}
\label{tab:slots_mwoz}
\end{table*}


\begin{table*}
\centering
\begin{tabular}{lccccccc}
\hline
& \textbf{attraction} & \textbf{hospital} & \textbf{hotel} & \textbf{profile} & \textbf{restaurant} & \textbf{taxi} & \textbf{train} \\ \hline
area & \checkmark &  & \checkmark &  & \checkmark &  &  \\ 
arriveby &  &  &  &  &  & \checkmark & \checkmark \\ 
day &  &  & \checkmark &  & \checkmark &  & \checkmark \\ 
department &  & \checkmark &  &  &  &  &  \\ 
departure &  &  &  &  &  & \checkmark & \checkmark \\ 
destination &  &  &  &  &  & \checkmark & \checkmark \\ 
email &  &  &  & \checkmark &  &  &  \\ 
food &  &  &  &  & \checkmark &  &  \\ 
idnumber &  &  &  & \checkmark &  &  &  \\ 
internet &  &  & \checkmark &  &  &  &  \\ 
leaveat &  &  &  &  &  & \checkmark & \checkmark \\ 
name & \checkmark &  & \checkmark & \checkmark & \checkmark &  &  \\ 
parking &  &  & \checkmark &  &  &  &  \\ 
people &  &  & \checkmark &  & \checkmark &  & \checkmark \\ 
phonenumber &  &  &  & \checkmark &  &  &  \\ 
platenumber &  &  &  & \checkmark &  &  &  \\ 
pricerange &  &  & \checkmark &  &  &  &  \\ 
stars &  &  & \checkmark &  &  &  &  \\ 
stay &  &  & \checkmark &  &  &  &  \\ 
time &  &  &  &  & \checkmark &  &  \\ 
type & \checkmark &  & \checkmark &  &  &  &  \\ \hline
\end{tabular}
\caption{Domain-Slot Combinations for SpokenWOZ.}
\label{tab:slots_swoz}
\end{table*}

\subsection{Replacements}
The full list of synonym and encoding replacements for slots and values are shown in Listings \ref{fig:ssa_replacements} and \ref{fig:esa_replacements}.

\begin{figure*}[htbp]
\centering
\begin{lstlisting}[language=json, caption={Domain and slot synonym replacements for SSA.}, label={fig:ssa_replacements}]
{
    "slots": {
        "pricerange": ["cost_range", "expense_range", "price_level"],
        "type": ["category", "classification", "kind"],
        "parking": ["parking"],
        "day": ["day", "bookday"],
        "bookday": ["day", "bookday"],
        "people": ["guests", "individuals", "persons"],
        "bookpeople": ["guests", "individuals", "persons"],
        "stay": ["nights", "duration", "length_of_visit"],
        "bookstay": ["nights", "duration", "length_of_visit"],
        "internet": ["wifi", "online_access", "broadband"],
        "name": ["name"],
        "area": ["location", "region", "zone", "neighborhood", "district"],
        "stars": ["rating", "grade"],
        "arriveby": ["arrival_time", "scheduled_arrival"],
        "leaveat": ["departure_time"],
        "destination": ["arrival_point", "end_point", "final_stop"],
        "departure": ["origin", "start_point", "beginning_location"],
        "food": ["cuisine", "style", "type"],
        "time": ["reservation", "slot"],
        "booktime": ["reservation", "slot"],
        "department": ["section", "division", "unit"],
        "email": ["email_address"],
        "idnumber": ["ID", "identification_number"],
        "phone": ["phone", "cell_number", "mobile_number"],
        "phonenumber": ["phone", "cell_number", "mobile_number"],
        "platenumber": ["license_plate", "plate"],
        "address": ["street_number"],
        "postcode": ["zipcode", "postal_code"],
        "ref": ["booking_number", "confirmation_number", "reservation_number"],
        "entrancefee": ["entry_fee", "admission_fee"],
        "openhours": ["business_hours", "hours_of_operation", "schedule"]
    },
    "domains": {
        "hotel": ["lodging", "accommodation", "motel"],
        "train": ["rail"],
        "attraction": ["sight", "landmark", "tourist_spot"],
        "restaurant": ["eatery", "diner", "cafe", "bistro", "food_place"],
        "hospital": ["medical_center", "health_facility"],
        "taxi": ["cab", "car", "uber", "lyft"],
        "profile": ["user", "account"],
        "police": ["cops", "law_enforcement"]
    }
}
\end{lstlisting}
\end{figure*}

\begin{figure*}[htbp]
\centering
\begin{lstlisting}[language=json, caption={Domain and slot synonym replacements for ESA.}, label={fig:esa_replacements}]
{
    "slots": {
        "pricerange": ["slot010", "slot011", "slot012", "slot013", "slot014"]
        "type": ["slot020", "slot021", "slot022", "slot023", "slot024"]
        "parking": ["slot030", "slot031", "slot032", "slot033", "slot034"]
        "day": ["slot040", "slot041", "slot042", "slot043", "slot044"]
        "bookday": ["slot050", "slot051", "slot052", "slot053", "slot054"]
        "people": ["slot060", "slot061", "slot062", "slot063", "slot064"],
        "bookpeople": ["slot070", "slot071", "slot072", "slot073", "slot074"],
        "stay": ["slot080", "slot081", "slot082", "slot083", "slot084"],
        "bookstay": ["slot090", "slot091", "slot092", "slot093", "slot094"],
        "internet": ["slot100", "slot101", "slot102", "slot103", "slot104"],
        "name": ["slot110", "slot111", "slot112", "slot113", "slot114"],
        "area": ["slot120", "slot121", "slot122", "slot123", "slot124"],
        "stars": ["slot130", "slot131", "slot132", "slot133", "slot134"],
        "arriveby": ["slot140", "slot141", "slot142", "slot143", "slot144"],
        "leaveat": ["slot150", "slot151", "slot152", "slot153", "slot154"],
        "destination": ["slot160", "slot161", "slot162", "slot163", "slot164"],
        "departure": ["slot170", "slot171", "slot172", "slot173", "slot174"],
        "food": ["slot180", "slot181", "slot182", "slot183", "slot184"],
        "time": ["slot190", "slot191", "slot192", "slot193", "slot194"],
        "booktime": ["slot200", "slot201", "slot202", "slot203", "slot204"],
        "department": ["slot210", "slot211", "slot212", "slot213", "slot214"],
        "email": ["slot220", "slot221", "slot222", "slot223", "slot224"],
        "idnumber": ["slot230", "slot231", "slot232", "slot233", "slot234"],
        "phone": ["slot240", "slot241", "slot242", "slot243", "slot244"],
        "phonenumber": ["slot250", "slot251", "slot252", "slot253", "slot254"],
        "platenumber": ["slot260", "slot261", "slot262", "slot263", "slot264"],
        "address": ["slot270", "slot271", "slot272", "slot273", "slot274"],
        "postcode": ["slot280", "slot281", "slot282", "slot283", "slot284"],
        "ref": ["slot290", "slot291", "slot292", "slot293", "slot294"],
        "entrancefee": ["slot300", "slot301", "slot302", "slot303", "slot304"],
        "openhours": ["slot310", "slot311", "slot312", "slot313", "slot314"]
    },
    "domains": {
        "attraction": ["domain10", "domain11", "domain12", "domain13", "domain14"],
        "hotel": ["domain20", "domain21", "domain22", "domain23", "domain24"],
        "hospital": ["domain30", "domain31", "domain32", "domain33", "domain34"],
        "police": ["domain40", "domain41", "domain42", "domain43", "domain44"],
        "profile": ["domain50", "domain51", "domain52", "domain53", "domain54"],
        "restaurant": ["domain60", "domain61", "domain62", "domain63", "domain64"],
        "taxi": ["domain70", "domain71", "domain72", "domain73", "domain74"],
        "train": ["domain80", "domain81", "domain82", "domain83", "domain84"]
    }
}
\end{lstlisting}
\end{figure*}

\subsection{Schema Augmentation Examples}
Full examples of the original prompt, SSA prompt, and ESA prompt for an example from MultiWOZ 2.2 are shown in Listings \ref{lst:schema_orig}, \ref{lst:schema_ssa}, and \ref{lst:schema_esa}.

\begin{figure*}[htbp]
\centering
\begin{lstlisting}[style=mypython, caption={Original data sample schema}, label={lst:schema_orig}]
### Schema:
- Slot: hotel-pricerange; Description: price budget of the hotel; Possible values: ['expensive', 'cheap', 'moderate']
- Slot: hotel-type; Description: what is the type of the hotel; Possible values: ['guesthouse', 'hotel']
- Slot: hotel-parking; Description: whether the hotel has parking; Possible values: ['free', 'no', 'yes']
- Slot: hotel-bookday; Description: day of the hotel booking; Possible values: ['monday', 'tuesday', 'wednesday', 'thursday', 'friday', 'saturday', 'sunday']
- Slot: hotel-bookpeople; Description: number of people for the hotel booking; Possible values: ['1', '2', '3', '4', '5', '6', '7', '8']
- Slot: hotel-bookstay; Description: length of stay at the hotel; Possible values: ['1', '2', '3', '4', '5', '6', '7', '8']
- Slot: hotel-stars; Description: star rating of the hotel; Possible values: ['0', '1', '2', '3', '4', '5']
- Slot: hotel-internet; Description: whether the hotel has internet; Possible values: ['free', 'no', 'yes']
- Slot: hotel-name; Description: name of the hotel; Possible values: []
- Slot: hotel-area; Description: area or place of the hotel; Possible values: ['centre', 'east', 'north', 'south', 'west']
- Slot: hotel-address; Description: address of the hotel; Possible values: []
- Slot: hotel-phone; Description: phone number of the hotel; Possible values: []
- Slot: hotel-postcode; Description: postal code of the hotel; Possible values: []
- Slot: hotel-ref; Description: reference number of the hotel booking; Possible values: []
- Slot: restaurant-pricerange; Description: price budget for the restaurant; Possible values: ['cheap', 'expensive', 'moderate']
- Slot: restaurant-area; Description: area or place of the restaurant; Possible values: ['centre', 'east', 'north', 'south', 'west']
- Slot: restaurant-food; Description: the cuisine of the restaurant you are looking for; Possible values: []
- Slot: restaurant-name; Description: name of the restaurant; Possible values: []
- Slot: restaurant-bookday; Description: day of the restaurant booking; Possible values: ['monday', 'tuesday', 'wednesday', 'thursday', 'friday', 'saturday', 'sunday']
- Slot: restaurant-bookpeople; Description: how many people for the restaurant reservation; Possible values: ['1', '2', '3', '4', '5', '6', '7', '8']
- Slot: restaurant-booktime; Description: time of the restaurant booking; Possible values: []
- Slot: restaurant-address; Description: address of the restaurant; Possible values: []
- Slot: restaurant-phone; Description: phone number of the restaurant; Possible values: []
- Slot: restaurant-postcode; Description: postal code of the restaurant; Possible values: []
- Slot: restaurant-ref; Description: reference number of the restaurant booking; Possible values: []
\end{lstlisting}
\end{figure*}

\begin{figure*}[htbp]
\centering
\begin{lstlisting}[style=mypython, caption={SSA schema}, label={lst:schema_ssa}]
### Schema:
- Slot: lodging-cost_range; Description: price budget of the hotel; Possible values: ['expensive', 'cheap', 'moderate']
- Slot: lodging-category; Description: what is the type of the hotel; Possible values: ['guesthouse', 'hotel']
- Slot: lodging-parking; Description: whether the hotel has parking; Possible values: ['free', 'no', 'yes']
- Slot: lodging-day; Description: day of the hotel booking; Possible values: ['monday', 'tuesday', 'wednesday', 'thursday', 'friday', 'saturday', 'sunday']
- Slot: lodging-guests; Description: number of people for the hotel booking; Possible values: ['1', '2', '3', '4', '5', '6', '7', '8']
- Slot: lodging-nights; Description: length of stay at the hotel; Possible values: ['1', '2', '3', '4', '5', '6', '7', '8']
- Slot: lodging-rating; Description: star rating of the hotel; Possible values: ['0', '1', '2', '3', '4', '5']
- Slot: lodging-wifi; Description: whether the hotel has internet; Possible values: ['free', 'no', 'yes']
- Slot: lodging-name; Description: name of the hotel; Possible values: []
- Slot: lodging-location; Description: area or place of the hotel; Possible values: ['centre', 'east', 'north', 'south', 'west']
- Slot: lodging-street_number; Description: address of the hotel; Possible values: []
- Slot: lodging-phone; Description: phone number of the hotel; Possible values: []
- Slot: lodging-zipcode; Description: postal code of the hotel; Possible values: []
- Slot: lodging-booking_number; Description: reference number of the hotel booking; Possible values: []
- Slot: eatery-cost_range; Description: price budget for the restaurant; Possible values: ['cheap', 'expensive', 'moderate']
- Slot: eatery-location; Description: area or place of the restaurant; Possible values: ['centre', 'east', 'north', 'south', 'west']
- Slot: eatery-cuisine; Description: the cuisine of the restaurant you are looking for; Possible values: []
- Slot: eatery-name; Description: name of the restaurant; Possible values: []
- Slot: eatery-day; Description: day of the restaurant booking; Possible values: ['monday', 'tuesday', 'wednesday', 'thursday', 'friday', 'saturday', 'sunday']
- Slot: eatery-guests; Description: how many people for the restaurant reservation; Possible values: ['1', '2', '3', '4', '5', '6', '7', '8']
- Slot: eatery-reservation; Description: time of the restaurant booking; Possible values: []
- Slot: eatery-street_number; Description: address of the restaurant; Possible values: []
- Slot: eatery-phone; Description: phone number of the restaurant; Possible values: []
- Slot: eatery-zipcode; Description: postal code of the restaurant; Possible values: []
- Slot: eatery-booking_number; Description: reference number of the restaurant booking; Possible values: []
\end{lstlisting}
\end{figure*}

\begin{figure*}[htbp]
\centering
\begin{lstlisting}[style=mypython, caption={ESA schema}, label={lst:schema_esa}]
### Schema:
- Slot: domain2-slot1; Description: price budget of the hotel; Possible values: ['expensive', 'cheap', 'moderate']
- Slot: domain2-slot2; Description: what is the type of the hotel; Possible values: ['guesthouse', 'hotel']
- Slot: domain2-slot3; Description: whether the hotel has parking; Possible values: ['free', 'no', 'yes']
- Slot: domain2-slot5; Description: day of the hotel booking; Possible values: ['monday', 'tuesday', 'wednesday', 'thursday', 'friday', 'saturday', 'sunday']
- Slot: domain2-slot7; Description: number of people for the hotel booking; Possible values: ['1', '2', '3', '4', '5', '6', '7', '8']
- Slot: domain2-slot9; Description: length of stay at the hotel; Possible values: ['1', '2', '3', '4', '5', '6', '7', '8']
- Slot: domain2-slot13; Description: star rating of the hotel; Possible values: ['0', '1', '2', '3', '4', '5']
- Slot: domain2-slot10; Description: whether the hotel has internet; Possible values: ['free', 'no', 'yes']
- Slot: domain2-slot11; Description: name of the hotel; Possible values: []
- Slot: domain2-slot12; Description: area or place of the hotel; Possible values: ['centre', 'east', 'north', 'south', 'west']
- Slot: domain2-slot27; Description: address of the hotel; Possible values: []
- Slot: domain2-slot24; Description: phone number of the hotel; Possible values: []
- Slot: domain2-slot28; Description: postal code of the hotel; Possible values: []
- Slot: domain2-slot29; Description: reference number of the hotel booking; Possible values: []
- Slot: domain6-slot1; Description: price budget for the restaurant; Possible values: ['cheap', 'expensive', 'moderate']
- Slot: domain6-slot12; Description: area or place of the restaurant; Possible values: ['centre', 'east', 'north', 'south', 'west']
- Slot: domain6-slot18; Description: the cuisine of the restaurant you are looking for; Possible values: []
- Slot: domain6-slot11; Description: name of the restaurant; Possible values: []
- Slot: domain6-slot5; Description: day of the restaurant booking; Possible values: ['monday', 'tuesday', 'wednesday', 'thursday', 'friday', 'saturday', 'sunday']
- Slot: domain6-slot7; Description: how many people for the restaurant reservation; Possible values: ['1', '2', '3', '4', '5', '6', '7', '8']
- Slot: domain6-slot20; Description: time of the restaurant booking; Possible values: []
- Slot: domain6-slot27; Description: address of the restaurant; Possible values: []
- Slot: domain6-slot24; Description: phone number of the restaurant; Possible values: []
- Slot: domain6-slot28; Description: postal code of the restaurant; Possible values: []
- Slot: domain6-slot29; Description: reference number of the restaurant booking; Possible values: []
\end{lstlisting}
\end{figure*}

\begin{figure*}[t]
\centering
\begin{lstlisting}[style=mypython, caption={Prompt Example}, label={lst:prompt_example}]
### Instructions: Give the dialogue state at the end of the given dialogue, formatted in JSON. Follow the schema and only use the given pre-defined slots and their possible values. `Possible values: []` means open-ended (the slot can take on any value). Omit any slots with empty values from your answer. If no slots can be filled from the dialogue, respond with an empty json object.

### Schema:
- Slot: restaurant-pricerange; Description: price budget for the restaurant; Possible values: ['cheap', 'expensive', 'moderate']
- Slot: restaurant-area; Description: area or place of the restaurant; Possible values: ['centre', 'east', 'north', 'south', 'west']
- Slot: restaurant-food; Description: the cuisine of the restaurant you are looking for; Possible values: []
- Slot: restaurant-name; Description: name of the restaurant; Possible values: []
- Slot: restaurant-bookday; Description: day of the restaurant booking; Possible values: ['monday', 'tuesday', 'wednesday', 'thursday', 'friday', 'saturday', 'sunday']
- Slot: restaurant-bookpeople; Description: how many people for the restaurant reservation; Possible values: ['1', '2', '3', '4', '5', '6', '7', '8']
- Slot: restaurant-booktime; Description: time of the restaurant booking; Possible values: []
- Slot: restaurant-address; Description: address of the restaurant; Possible values: []
- Slot: restaurant-phone; Description: phone number of the restaurant; Possible values: []
- Slot: restaurant-postcode; Description: postal code of the restaurant; Possible values: []
- Slot: restaurant-ref; Description: reference number of the restaurant booking; Possible values: []
- Slot: attraction-area; Description: area to search for attractions; Possible values: ['centre', 'east', 'north', 'south', 'west']
- Slot: attraction-name; Description: name of the attraction; Possible values: []
- Slot: attraction-type; Description: type of the attraction; Possible values: ['architecture', 'boat', 'cinema', 'college', 'concerthall', 'entertainment', 'museum', 'multiple sports', 'nightclub', 'park', 'swimmingpool', 'theatre']
- Slot: attraction-entrancefee; Description: how much is the entrance fee; Possible values: []
- Slot: attraction-openhours; Description: open hours of the attraction; Possible values: []
- Slot: attraction-address; Description: address of the attraction; Possible values: []
- Slot: attraction-phone; Description: phone number of the attraction; Possible values: []
- Slot: attraction-postcode; Description: postal code of the attraction; Possible values: []


### Dialogue:
USER: i am looking for a college type attraction .
SYSTEM: there are 18 colleges i have found , would you prefer 1 in town centre or in the west ?
USER: i would like to visit on in town centre please .
\end{lstlisting}
\end{figure*}

\subsection{Licenses}
MultiWoz is available under an MIT License, and SpokenWoz is available under a CC-BY-NC 4.0 license. Gemma-2 is available pursuant to the \href{https://ai.google.dev/gemma/terms}{Gemma Terms of Use}. Mistral is available under an Apache 2.0 license. All models and datasets are intended for research purposes, which is consistent with this work. 

 Datasets are widely used and reported not to contain personal information or offensive content.

\end{document}